%% file: 0-main.tex
\documentclass[conference]{IEEEtran}
\IEEEoverridecommandlockouts
\usepackage{amsmath,amssymb,amsfonts}
\usepackage{algorithmic}
\usepackage{graphicx}
\usepackage{textcomp}
\usepackage{xcolor, xspace}
\usepackage[colorlinks,citecolor=blue,urlcolor=blue,linkcolor=blue,bookmarks=false,hypertexnames=true]{hyperref}
\usepackage{booktabs, subcaption, array, tikz, pgfplots, comment}
\usepackage{etoolbox}
\usepackage{subcaption, multirow}
\usepackage[utf8]{inputenc}
\DeclareUnicodeCharacter{1EF3}{\`y}

\usepackage[
  sorting=none,
  style=ieee, 
  backend=biber,
  isbn=false,
  url=true,
  doi=false,
  url=false,
  mincrossrefs=100,
  maxnames=12,
  firstinits=true
]{biblatex}

\AtEveryBibitem{\clearname{editor}}

\newif\ifshowcomments
 \showcommentstrue
\ifshowcomments
\newcommand{\mynote}[2]{\fbox{\bfseries\sffamily\scriptsize{#1}}
{\small$\blacktriangleright$\textsf{\emph{#2}}$\blacktriangleleft$}}
\else
\newcommand{\mynote}[2]{}
\fi

\newcommand{\algo}{\texttt{AGIC}\xspace}

\usepackage{flushend}

\makeatletter
\patchcmd{\@makecaption}
{\scshape}
{}
{}
{}
\makeatother
\def\BibTeX{{\rm B\kern-.05em{\sc i\kern-.025em b}\kern-.08em
    T\kern-.1667em\lower.7ex\hbox{E}\kern-.125emX}}
\begin{document}

\title{\algo: Approximate Gradient Inversion Attack on Federated Learning\\
}

\author{\IEEEauthorblockN{Jin Xu, Chi Hong, Jiyue Huang, Lydia Y. Chen, Jérémie Decouchant}
\IEEEauthorblockA{\textit{Delft University of Technology, The Netherlands}\\
j.xu-21@student.tudelft.nl, \{c.hong, j.huang-4, y.chen-10, j.decouchant\}@tudelft.nl}
}

\maketitle

\thispagestyle{plain}
\pagestyle{plain}

\begin{abstract}
Federated learning is a private-by-design distributed learning paradigm where clients train local models on their own data before a central server aggregates their local updates to compute a global model.  Depending on the aggregation method used, the local updates are either the gradients or the weights of local learning models, e.g., FedAvg aggregates model weights.
Unfortunately, recent reconstruction attacks apply a gradient inversion optimization on the gradient update of a single mini-batch to reconstruct the private data used by clients during training.
As the state-of-the-art reconstruction attacks solely focus on single update, realistic adversarial scenarios are overlooked, such as 
observation across multiple updates and updates trained from multiple mini-batches.
A few studies consider a more challenging adversarial scenario where only model updates {based on multiple mini-batches} are observable, and resort to computationally expensive simulation to untangle the 
underlying samples for each local step. 
In this paper, we propose \algo, a novel \textbf{A}pproximate \textbf{G}radient \textbf{I}nversion Atta\textbf{c}k that efficiently and effectively reconstructs images from both model or gradient updates, and across multiple epochs.
In a nutshell, \algo (i) approximates gradient updates of used training samples from model updates to avoid costly simulation procedures,
(ii) leverages gradient/model updates collected from multiple epochs, and (iii) assigns increasing weights to layers with respect to the neural network structure for reconstruction quality. 
We extensively evaluate \algo on three datasets, namely CIFAR-10, CIFAR-100 and ImageNet. Our results show that \algo increases the peak signal-to-noise ratio (PSNR) by up to 50\% compared to two representative state-of-the-art gradient inversion attacks.
Furthermore, \algo is faster than the state-of-the-art simulation-based attack, e.g., it is 5x faster when attacking FedAvg with 8 local steps in between model updates.

\let\thefootnote\relax\footnotetext{
	\fbox{\parbox{\dimexpr.9\columnwidth\fboxsep-2\fboxrule\relax}{This document is a preprint of a paper accepted at the 41st International Symposium on Reliable Distributed Systems (SRDS 2022).}}
}
\end{abstract}


\input{1-introduction}
\input{2-background}
\input{3-Algorithm}
\input{4-Evaluation}
\input{5-related}

\section{Conclusion}
\label{sec:conclusion}

In this paper, we presented \algo, an accurate and fast gradient inversion attack that can leverage both model updates and gradient updates across multiple epochs. \algo uses a one-batch approximation to convert model updates into approximate gradient updates of a larger mini-batch. \algo also leverages updates from multiple epochs with update matching to jointly reconstruct specific training samples accurately. Finally, \algo assigns different weights to layers in the objective function, which significantly improves the reconstruction quality. Compared to previous works, \algo can be used with more general federating learning system settings. Our experiments demonstrate that \algo reconstructs samples with up to 50\% PSNR improvement compared to state-of-the-art baselines, and is up to 5x faster than simulation-based attacks, which is the only baseline that can attack FedAvg.

\printbibliography

\end{document}

%% file: 1-introduction.tex
\section{Introduction} 

Federated learning (FL)~\cite{bonawitz2019towards, bonawitz2017practical, konevcny2016federated, mcmahan2017communication, mcmahan2016federated} is a popular collaborative learning paradigm that aims at providing accurate predictive models while preserving the clients' data privacy and reducing communication costs. In a FL system, the training data of a client never leaves its initial premises. In each global round, the server first sends the most recent model
to clients, which then train the local model on their private data and send gradients or model updates back to the server. At the end of a global round, the server is able to update the global model by aggregating all the gradients or model updates it has received. 

An attacker that would compromise the server would observe model parameters and their updates, but would not have access to training samples. Therefore, FL is often assumed to safely protect the clients' local data. However, recent works have shown that the observation of model parameters and gradient updates might allow attributes of local samples to be leaked~\cite{melis2019exploiting}, class representatives to be inferred~\cite{hitaj2017deepgan, wang2019beyond}, and even real training samples to be reconstructed~\cite{zhu2020deepdlg, geiping2020inverting}. 

Gradient inversion attacks that directly reconstruct training samples based on a model and gradient updates on it result in the most serious data leakages. Most gradient inversion attacks are optimization-based. In an optimization-based gradient inversion attack, the adversary randomly initializes dummy samples, and executes forward and backward propagation on them to obtain dummy gradients. The dummy samples are then optimized to minimize the sum of the distance between the observed real gradients and the dummy gradients, and regularization terms. To improve the reconstruction performance, some works define new distance functions and regularization terms~\cite{geiping2020inverting}, while others exploit prior knowledge, e.g., by using batch normalization data~\cite{yin2021seegradinversion} or pre-trained generative models~\cite{jeon2021gradientgen}.

Gradient inversion attacks are designed to reconstruct samples from gradient updates. However, in practice, federated learning systems often use federated averaging (FedAvg)~\cite{mcmahan2017communication} where clients send model updates after conducting multiple local steps, each executed over a mini-batch, in order to further reduce the communication overhead. Few previous works have discussed how to attack model updates generated by FedAvg. These works either use simulation~\cite{geiping2020inverting}, which is slow and incompatible with label inference, a sub-task that significantly improves the reconstruction performance~\cite{zhao2020idlg}, or 
make an averaging approximation~\cite{geng2021general} to estimate the gradient of a full batch for FedAvg, which does not extend to the most general case where mini-batches are used. 

In this paper, we describe \algo, a novel Approximate Gradient Inversion Attack that specifically targets federated learning systems based on FedAvg's model updates, and is also compatible with gradient updates. Overall, \algo makes the following contributions.

First, to avoid the computational overhead of simulation attacks, \algo leverages a one-batch approximation, which assimilates multiple local steps executed over multiple mini-batches as a single local step executed over a single aggregated mini-batch. The corresponding gradients of that larger mini-batch can also be approximated based on the received model update.  
Thanks to this one-batch approximation, \algo is faster than previous simulation-based attacks on FedAvg and is compatible with label inference. 

Second, based on the observation that the attacker is able to observe updates across multiple epochs, \algo better exploits the information contained in multiple updates that have been computed over common data samples. \algo matches the collected updates with specific training samples and jointly optimizes with the updates to improve the reconstruction quality of the specific training samples. 

Finally, \algo assigns different weights to the gradients of different layers in the distance function to optimize, inspired by the results of recent works on model compression that have shown that model layers have unequal effects on model accuracy and other metrics~\cite{yang2017designing, elkerdawy2020filter}. Interestingly, \algo uses a layer weight modifier for convolutional neural networks (CNN) that use ReLU as activation function, to balance the contribution of each convolution layer, because ReLU imports zeros into gradients.

Our experiments with three datasets show that \algo outperforms two representative state-of-the-art baselines in both reconstruction quality and efficiency. When attacking gradient updates, \algo's peak signal-to-noise ratio is up to 50\% higher than with both baselines.
When attacking FedAvg model updates configured with 8 local steps, \algo reconstructs samples five times faster than the only applicable  baseline, a state-of-the-art simulation-based attack. 

The remainder of this paper is organized as follows. Section~\ref{sec:background} provides some background on federated learning and gradient inversion attacks. Section~\ref{sec:overview} introduces our system and threat models, and an overview of \algo. Section~\ref{sec:fedavg} describes \algo's three key features. Section~\ref{sec:experiments} evaluates \algo's performance. Finally, Section~\ref{sec:conclusion} concludes this paper. 

%% file: 2-background.tex
\section{Background} \label{sec:background}

\subsection{Federated learning} 
\label{subsec:bg_fl}

Federated learning (FL)~\cite{mcmahan2016federated, bonawitz2017practical} is a collaborative learning paradigm that allows distributed data owners (called clients) to jointly learn machine learning models without centrally pooling their data. 
Clients train the model using their own local data and rely on a central server to aggregate their learning results. The FL system progresses over multiple global rounds. At the beginning of a global round, the server sends the latest global model to clients. Clients then train the latest model on their own data, and send their training results, i.e., gradients or model updates, back to the server.

The local training process typically follows the stochastic gradient descent (SGD) algorithm, which iterates through the data using batches. A batch might contain the full local dataset, but in the more general case a batch is a subset of the local dataset and is also called a mini-batch.
During an \textit{epoch}, clients iterate over their complete training data once. A client may use its full dataset in a single batch during one round, which is then regarded as an epoch.
We focus on the more general mini-batch case, where the full dataset is split into multiple mini-batches, and the client iterates over these mini-batches in multiple global rounds 
during an epoch.

\begin{figure}[tp!]
\centering
	\includegraphics[width=\columnwidth]{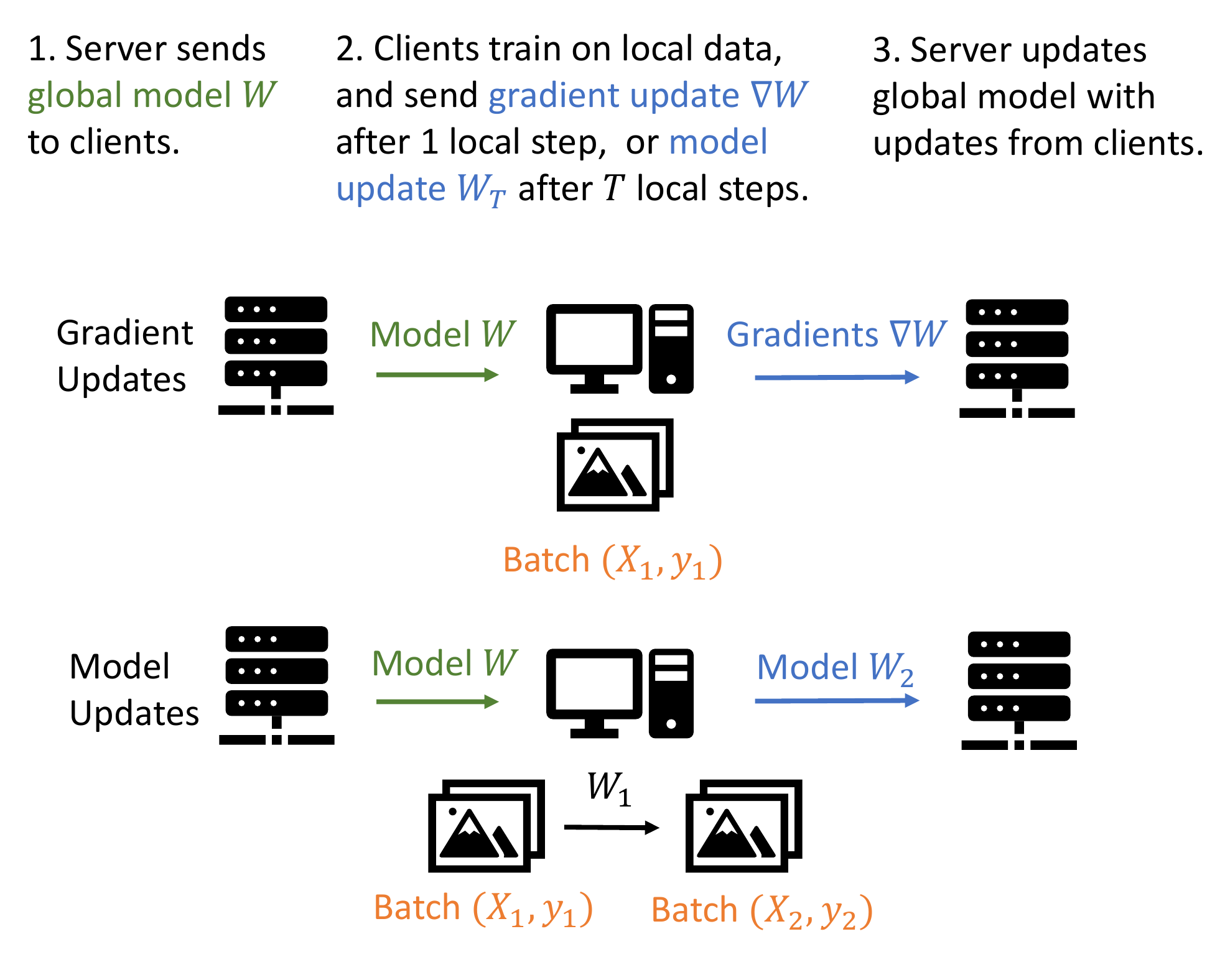}
	\caption{A global federated learning round with gradient update and with model update.}
	\label{fig:fl2}
\end{figure}

Figure~\ref{fig:fl2} depicts two typical local training procedures of FL learning systems that respectively send gradient updates after one local step (top), or send model updates after multiple local steps (bottom), which is the case  with federated averaging  (FedAvg)~\cite{mcmahan2017communication}. 
With the first method, the clients execute a single local training step on a batch of local samples and their labels $(X, y)$ with the latest received global model $W$, and send back the corresponding gradient update $\nabla W$. With FedAvg, the clients run $T$ local steps on $T$ batches, which means that they will update their local model parameters $T$ times, and then send the trained model weights $W_T$ back to the server. In practical scenarios, FedAvg is more commonly used because it reduces the amount of data transmitted over the network, in particular because a client can process several batches during a global round, i.e., before sending its results to the server. 
In the rest of this paper, we will note $\Delta W {=} W_T {-} W$, the difference between the updated model sent by the client at the end of a round and the global model it received at its beginning. 

\begin{figure*}[tp!]
	\centering
	\begin{subfigure}[t]{0.48\textwidth}
		\includegraphics[scale=0.52]{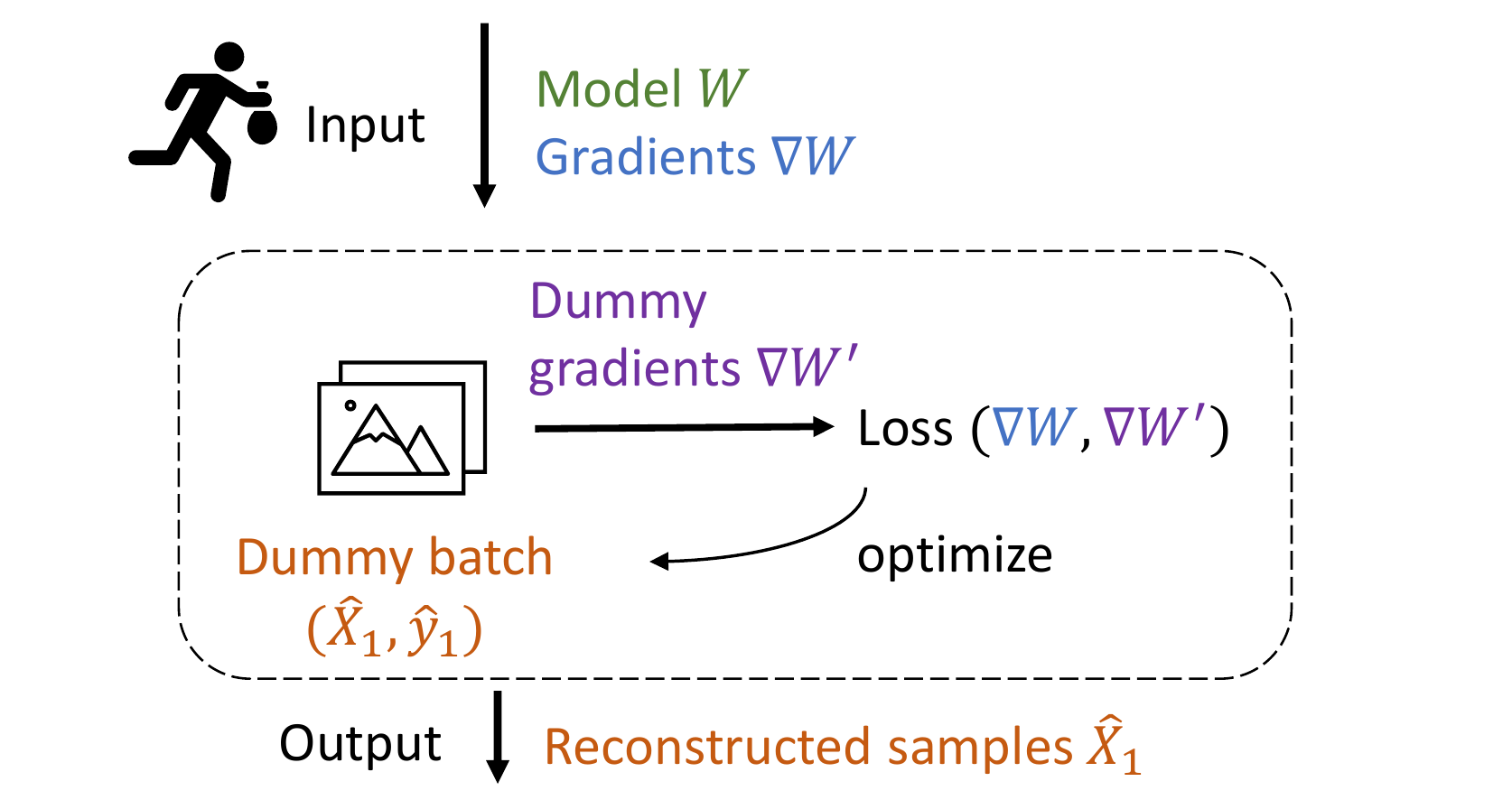}
		\subcaption{Attack based on gradient updates.}
		\label{fig:attack2}
	\end{subfigure}
	\begin{subfigure}[t]{0.48\textwidth}
		\includegraphics[scale=0.52]{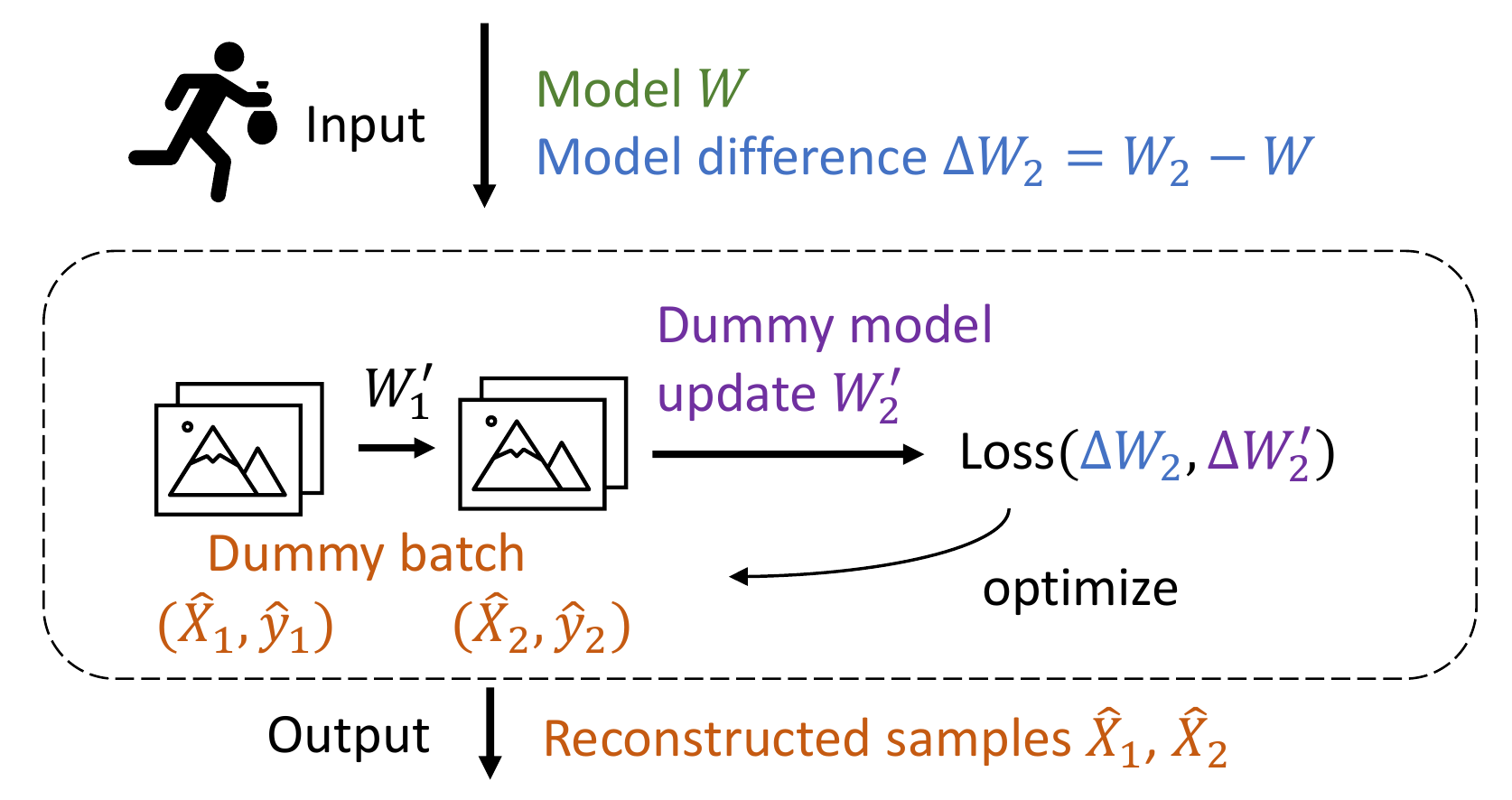}
		\subcaption{Simulation-based attack based on model updates with 2 local steps using simulation. }
		\label{fig:attack3}
	\end{subfigure}
	\caption{Gradient inversion attacks based on gradient updates and model updates.}
	\label{fig:attack1}
\end{figure*}

\subsection{Gradient inversion attacks} 
\label{subsec:bg_attack}

Even though the clients and the server only exchange intermediate learning results such as models or gradient updates, it has been shown that sensitive information from local training datasets can be inferred in FL systems~\cite{melis2019exploiting, wang2019beyond}. Furthermore, it has been shown that an attacker that obtains gradient updates can launch a gradient inversion attack to recover the training samples that were used by clients to generate the gradients~\cite{zhu2020deepdlg}. Gradient inversion attacks generally assume the attacker is an honest-but-curious server~\cite{geiping2020inverting}.

Most existing gradient inversion attacks solve an optimization problem, as illustrated in Figure~\ref{fig:attack2}. After retrieving gradient updates $\nabla W$, the attacker generates dummy samples $(\hat{X}, \hat{y})$ and minimizes the distance between the received gradients $\nabla W$ and its dummy gradients $\nabla W'$, which are retrieved by feeding dummy samples through the obtained model in one forward-backward pass. During the optimization process, the values of the dummy samples are optimized, so that at the end of the attack the dummy samples approximate the training samples.  

An optimization-based gradient inversion attack typically relies on an optimization objective such as: 
\begin{equation} 
\label{eq:attack}
\min_{\hat X} \textnormal{Dist}(\frac{\partial \mathcal{L}(F(W, \hat X), \hat y)}{\partial W}, \nabla W) +\textnormal{Reg}(\hat X, \hat y) 
\end{equation}

Formula~\ref{eq:attack} uses the following notations. $W$ is the current values of the trainable parameters of the attacked neural network. $\nabla W = \frac{\partial \mathcal{L}(F(W, X), y)}{\partial W}$ is the gradient update received from a client. $F$ is the forward propagation function of the model. $\mathcal{L}$ is the loss function of the model. $\hat X$ is the generated reconstruction samples and $\hat y$ is the labels of the samples.  \textnormal{Dist} is a distance function such as the L2 distance~\cite{zhu2020deepdlg} and the cosine distance~\cite{geiping2020inverting}, which are the two most widely used distance functions in gradient inversion attacks. \textnormal{Reg} refers to regularization terms. For attacks on image classification tasks, additional regularization terms can be leveraged to generate more natural images. For example, total variation~\cite{geiping2020inverting} is used to reduce image noise, and clipping terms~\cite{geng2021general} are used to prevent abnormal values that are out of range for a pixel. The sum of the distance and regularization terms for generated samples forms the objective to be minimized.

In the previous paragraph, we mention that both dummy samples $\hat{X}$ and dummy labels $\hat{y}$ are simultaneously optimized to recover the local training data $(X, y)$. In reality, to avoid jointly optimizing on the labels and reducing the complexity of the optimization problem, recent analytical approaches infer the labels before conducting the optimization by analyzing the distribution of the gradient tensor of the last fully connected layer~\cite{zhao2020idlg, yin2021seegradinversion}. Label inference is a crucial step to improve reconstruction quality of the most recent gradient inversion attacks.

Designing accurate gradient inversion attacks on model updates with mini-batches is more difficult than on gradient updates. Indeed, assuming a batch size $B$ for each local step, an adversary attacking gradient updates needs to reconstruct $B$ samples. With model updates, from $T$ local-step model updates respectively computed over $T$ different mini-batches of size $B$, the adversary needs to reconstruct $TB$ samples for an identical update size, i.e., number of model parameters. 

While most gradient inversion attacks can only be applied to gradient updates, a recent attack on model updates ~\cite{geiping2020inverting} simulates the execution of $T$ local steps in each optimization iteration, retrieves the dummy model update $W_T'$ after the last local step, and optimizes the dummy samples in all mini-batches $\hat X_1, ..., \hat X_T$ based on the difference between the dummy model and the observe model $\Delta W_T'=W_T'-W$ and the real observed model difference $\Delta W_T$. Figure~\ref{fig:attack3} illustrates a simulation-based gradient inversion attack based on FedAvg model updates.  

%% file: 3-Algorithm.tex
\section{Overview of \algo} \label{sec:overview}

This section defines our system and threat models, and provides an overview of \algo, our gradient inversion attack. 

\subsection{System and adversarial models}

We consider a FL system with one server that aggregates updates and multiple clients. 
We assume all clients to be correct and that the server is  honest-but-curious. Therefore, our adversary observes the communications between the server and the clients and can  retrieve both models and model updates. The attack can occur at any phase of the training process to target untrained, in-training or well-trained networks. The attacker may observe updates from different epochs during training. During an epoch a client iterates over its whole dataset, possibly using different mini-batches.  

We focus on attacking convolutional neural networks (CNN), which are used for image classification, one of the most popular uses of FL systems.
For example, it can be used by medical institutes to collaboratively train a diagnosis model~\cite{rieke2020future} without sharing patient-related information. 

\subsection{Attack overview}

To successfully run gradient inversion attacks on FedAvg, \algo relies on three key features: the one-batch approximation for federated averaging, the leveraging of model updates from multiple epochs, and the assignment of layer weights in the distance function.

\begin{figure}[t]
	\centering
	\includegraphics[width=\columnwidth]{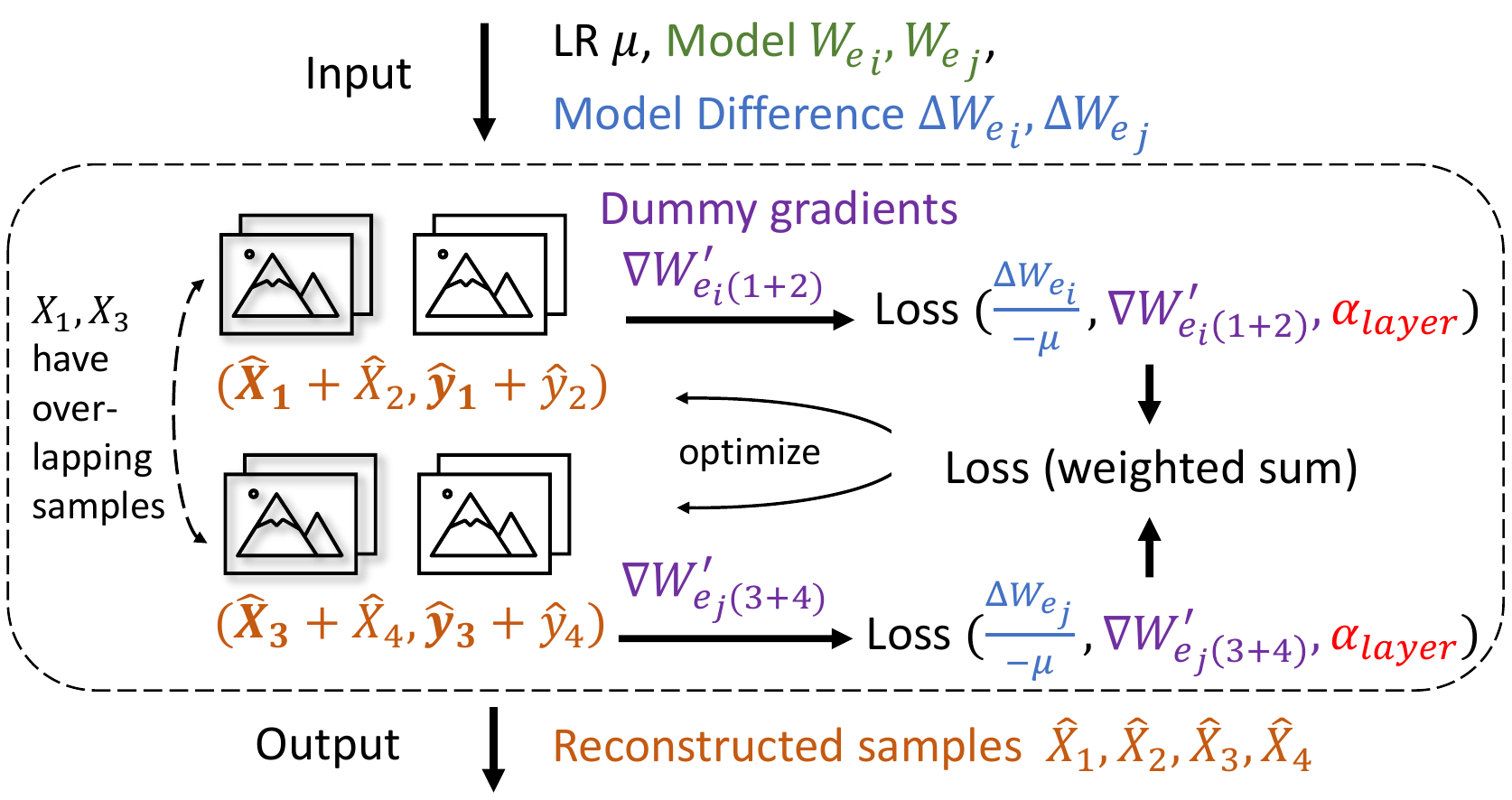}
	\caption{Overview of \algo.}
	\label{fig:overview}
\end{figure}

Figure~\ref{fig:overview} illustrates \algo and its key features.
Compared to the simulation attack on FedAvg, which was presented in Figure~\ref{fig:attack3}, this Figure shows \algo key features.  First, \algo uses a one-batch approximation so that mini-batches $\hat X_1, \hat X_2$ within a two-step FedAvg update compose a larger batch $\hat X_1 {+} \hat X_2$ (cf. \S\ref{sec:one-batch}).
Figure~\ref{fig:overview} also illustrates two FedAvg model updates from two different epochs $e_i, e_j$  that have used overlapping samples in their first local step, i.e., first mini-batch, respectively $X_1$ and $X_3$. Tensors $\hat X_1$ and $\hat X_3$ thus have overlapping components and can be jointly optimized. (cf. \S\ref{sec:matching}). $\alpha_{layer}$ represent the different layer weights that \algo uses in its distance function (cf. \S\ref{sec:weights}).

\textbf{Base loss function used during optimization.}
The base loss function that \algo uses to reconstruct images from an update is the negative cosine distance between the dummy gradients $\nabla W'=\frac{\partial \mathcal{L}(F(\hat X, W), y)}{\partial W}$ and the received real gradients $\nabla W$, to which is added  a total variation (TV) regularization term that reduces image noises with weight $\zeta_{TV}$~\cite{rudin1992nonlinear}:  

\begin{equation} \label{eq:invg}
\min_{\hat X} 1 -\cos (\nabla W', \nabla W) +\zeta_{TV} \textnormal{TV}(\hat X)
\end{equation}

\begin{equation} \label{eq:cos}
\cos(\nabla W', \nabla W) = \frac{\nabla W' \cdot \nabla W}{\parallel \nabla W' \parallel_2  \ \parallel \nabla W \parallel_2}
\end{equation}

\noindent The dummy samples are optimized based on the gradient values from back propagation of the loss. 
\algo modifies this base loss function to adapt to different attack scenarios. 

\algo is a novel approximate gradient inversion attack that is applicable to both model updates and gradient updates.
We focus on the more challenging model update scenario, but our methods (namely, leveraging multiple epochs and choosing layer weights) also apply to gradient updates. 

\section{Key implementation features of \algo} \label{sec:fedavg}

This section details \algo's three key features that enable its use with FedAvg and increase its accuracy: one-batch approximation, use of multiple updates, and layer weights.  

\subsection{One-batch approximation for FedAvg}
\label{sec:one-batch}

As previously explained, accurately reconstructing samples from FedAvg's model updates is more difficult than with gradient updates because a model update is computed using more samples. 
When attacking FedAvg with possibly multiple local steps, \algo is faster and more accurate than simulation-based gradient inversion attacks thanks to its one-batch approximation. This approximation allows \algo not to use computationally expensive simulation and to be compatible with label inference, as there is no need to assign inferred labels to different local steps. 

The one-batch approximation is equivalent to assuming that the model remains mostly identical after each local step. This assumption has been discussed in large mini-batch training~\cite{goyal2017accurate}. A similar first-order approximation has been used in meta learning~\cite{finn2017model}. It should be noted that if model parameters change drastically in each local step, e.g. when the learning rate of local SGD is $1\times 10^{-2}$, the one-batch approximation should not be used. In our experiments, with a learning rate of $1\times 10^{-4}$, the approximation brings good performance (cf. \ref{subsec:exp_overall}).

\begin{figure}[t]
	\centering
	\includegraphics[width=\columnwidth]{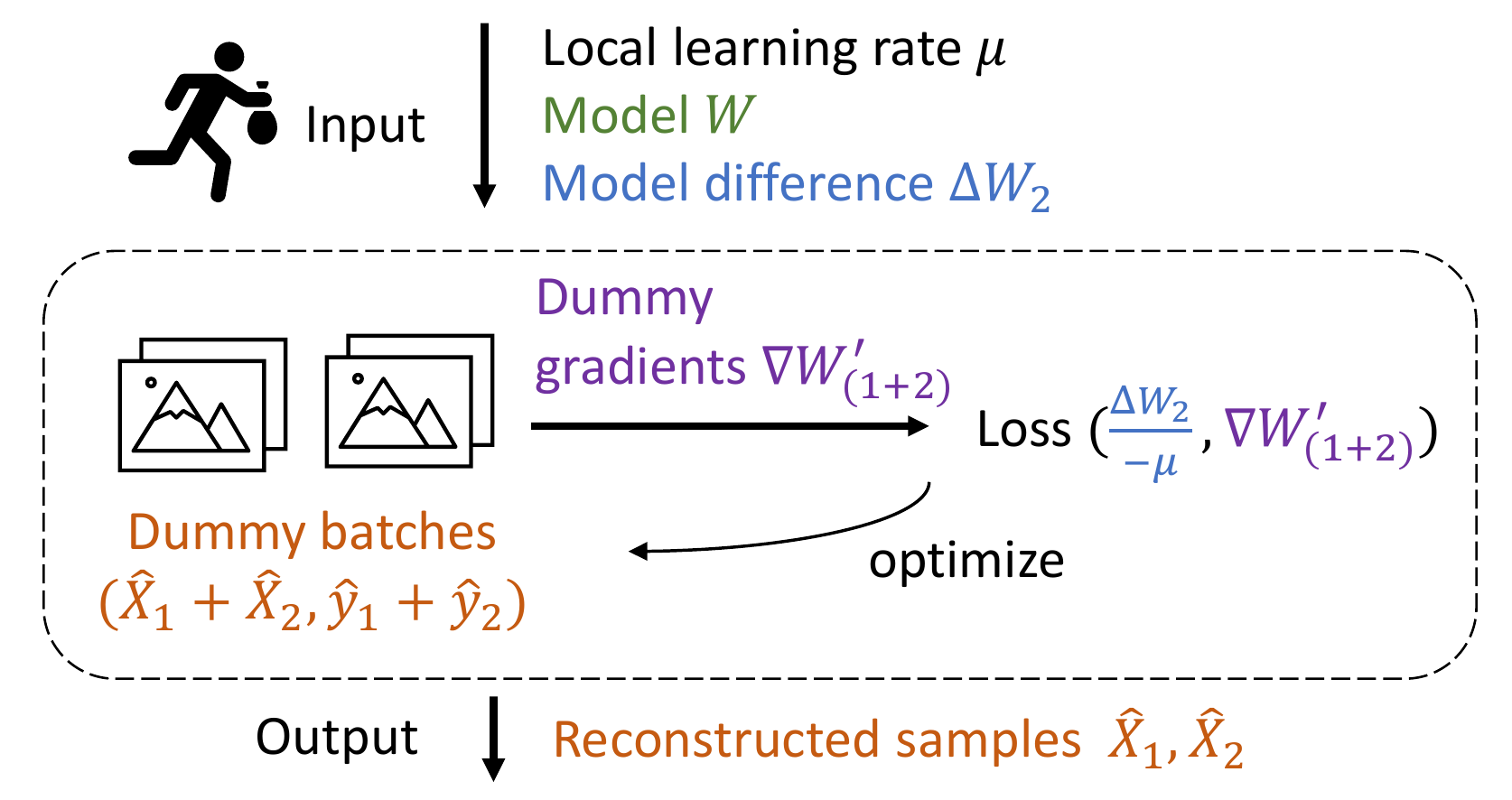}
	\caption{Gradient inversion attack with \algo's one-batch approximation on FedAvg model updates. Here, clients use 2 local steps before sharing model updates with the server.}
	\label{fig:flow2}
\end{figure}

Figure~\ref{fig:flow2} illustrates \algo's one-batch approximation on FedAvg's updates. Compared to the multi-step simulation of~\cite{geiping2020inverting}, which is presented in Figure~\ref{fig:attack3}, \algo's approximation only runs one local step with the aggregated batch, and its loss function is based on approximated gradients instead of models difference $\Delta W$. 
More precisely, given all the mini-batches used in one FedAvg update $X_1, ..., X_n$ and their inferred labels $y_1, ..., y_n$, and local learning rate $\mu$, \algo makes the approximation that all the mini-batches compose a larger batch $X_{A} = [X_1 , ... , X_n]$ to produce the received model update: 

\begin{equation} \label{eq:onebatch}
\Delta W \approx \sum_{j}^{N} -\mu\frac{\partial \mathcal{L}(F(W, X_j), y_j)}{\partial W} = -\mu\frac{\partial \mathcal{L}(F(W, X_A), y_A)}{\partial W}
\end{equation}

Based on Formula~\ref{eq:onebatch}, \algo computes the approximated gradients $\nabla W = \frac{\Delta W}{-\mu}$ for aggregated batch $X_A$ from the model updates given then local learning rate. Then \algo initializes dummy samples $\hat X_A$ as the size of $X_A$, and optimizes on the distance between dummy gradients from $\hat X_A$ and the computed gradients from received model updates. Because the objective function is the distance between gradients, any improvement method designed for gradient updates can be directly applied on FedAvg model updates with \algo 's approximation.

\begin{figure}[tp]
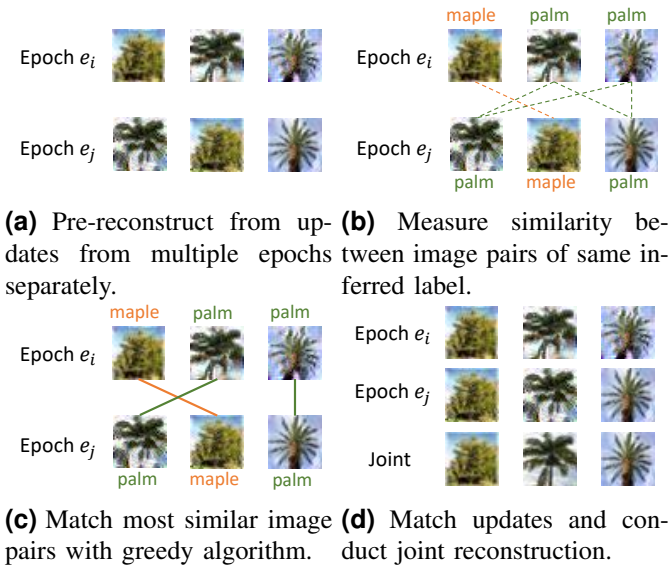

	\centering
	\begin{subfigure}[t]{0.24\textwidth}
		\centering
		\includegraphics[width=0.95\linewidth,page=2]{"match1.pdf"}
		\caption{Pre-reconstruct from updates from multiple epochs separately.} 
	\end{subfigure}
	\begin{subfigure}[t]{0.24\textwidth}
	\centering
	\includegraphics[width=0.95\linewidth,page=3]{"match1.pdf"}
	\caption{Measure similarity between image pairs of same inferred label.} 
\end{subfigure}
	\begin{subfigure}[t]{0.24\textwidth}
	\centering
	\includegraphics[width=0.95\linewidth,page=4]{"match1.pdf"}
	\caption{Match most similar image pairs with greedy algorithm.} 
\end{subfigure}
	\begin{subfigure}[t]{0.24\textwidth}
	\centering
	\includegraphics[width=0.95\linewidth,page=5]{"match1.pdf"}
	\caption{Match updates and conduct joint reconstruction.} 
\end{subfigure}
	\caption{Reconstruction from updates from multiple epochs. We use updates with batch size 1 for illustration purposes.}
	\label{fig:gradmatch}
\end{figure}

\subsection{Leveraging updates from multiple epochs}
\label{sec:matching}

A training sample is used in many global rounds, each round producing an update, and thus is involved in multiple updates that are communicated to the server. The attacker can therefore observe several updates that are based on some specific samples. \algo demonstrates that the adversary can better reconstruct the samples by making use of the updates it observes.

Prior work by Geng et al. has shown that reconstruction quality of a specific batch is improved if jointly optimizing with more pairs of model parameters $W_k$ and their corresponding gradient update from that batch $W'_k$~\cite{geng2021general}. However, this method only works on full batch gradient descent and does not provide a solution to the general case where the client uses multiple mini-batches to go over its data as in an epoch. In contrast, \algo is able to reconstruct training samples using multiple model updates or gradient updates from different mini-batches across epochs during training.

\algo processes updates from multiple epochs in two steps: (i) \textit{update matching}, which matches the collected updates whose corresponding samples are overlapping; and (ii) \textit{joint reconstruction}, which reconstructs the samples from the matching updates. We discuss those two steps in the following, and Figure~\ref{fig:gradmatch} illustrates them with a batch size equal to 1. 

{\bf Step 1: Update matching.} \algo first matches updates that have overlapping samples so that they can be jointly reconstructed afterwards. If the training process does not shuffle the samples used between epochs, or if the entire local dataset is used in a full batch, then the matching process is trivial. In the general case, a more complex algorithm is required to match gradient updates of single mini-batches or FedAvg model updates based on multiple mini-batches. \algo's matching mechanism first conducts a reconstruction process for each update to obtain a first reconstruction of images, and then identifies the most similar pairs of reconstructed images.

As the adversary observes updates from two complete epochs, the pre-reconstructed samples from the first epoch's updates can be mapped one-to-one with the samples from the updates of the second epoch. 
Equivalently said, for any sample $X_c$, two updates from the two epochs have been trained on it. 

To match pairs of reconstructed images, we evaluate the similarity of pairs of images by processing the images with an average pooling layer that takes average value of each small part of an image in order to mitigate noises, and computing their mean squared error. We refer the reader to \S\ref{subsec:exp_epoch}, which demonstrates the performance improvement that using a pooling layer brings. 

We use a simple greedy algorithm to find a one-to-one mapping between reconstructed images from two epochs. First, we measure the similarity of each image pair and sort them. From the highest similarity to the lowest one, we check each pair of still unmatched images and associate them to each other. Using this method, all pre-reconstructed images from one epoch can be matched to those from another epoch. For any matched image pairs, which are then considered to reconstruct the same sample, \algo executes a joint reconstruction using the all updates they participated in (Step 2). 

It is easy to extend this method to leverage more than two epochs, by successively computing matching update pairs from consecutive epoch pairs. 
A specific sample is then associated to the image pairs reconstructed from updates from consecutive epochs, and to one reconstructed image per epoch. A sample is therefore associated to exactly one update per epoch. 

Since label inference can be applied before optimizing for each update, \algo uses a label-based filter. Based on their inferred label, \algo computes the similarity between image pairs that have the same inferred label to reduce the search space and limit the possibility of incorrect matching, thereby increasing accuracy. For a multi-sample batch, generated either from gradient update directly or from the one-batch approximation of model update, label inference returns a list of labels for all samples.
Therefore, whether two samples can be matched depends on whether their inferred label lists share at least a common label. 

{\bf Step 2: Joint reconstruction with multiple updates.} \algo jointly optimizes updates that have been identified as having been trained over a common sample, in order to improve reconstruction quality. We use gradients during the optimization process, obtained either from the one-batch approximation of model update or directly from received gradient update. Our distance function uses the summation of distance of all pairs of input gradients and dummy gradients, with weights $\gamma_i$ assigned. A decreasing weight is assigned to the gradients generated later during the training process, because the reconstruction quality of gradient inversion attacks is higher in the earliest training phases. These gradient weights are set as experimental hyperparameters (\S\ref{sec:experiments} details the values we use). 

Let us assume that after the update matching step, \algo has identified that gradients $\nabla W_i$ with corresponding model parameters $W_i$ from different epochs have been trained over a given sample $X_c$. We note $\nabla W'_i$ the dummy gradients from the dummy batches. \algo's distance function with multiple gradient updates is the following.

\begin{equation} \label{eq:epoch1}
\min_{\hat X_k} \sum_{k} (1 - \frac{\nabla W'_k \cdot\nabla W_k}{\parallel \nabla W'_k \parallel_2 \ \parallel \nabla W_k \parallel_2}) \cdot \gamma_k
\end{equation}

The reconstruction results of \algo's joint reconstruction are of higher quality than those that would be obtained by separately optimizing based on single updates. To illustrate why, let us consider two gradient updates from two epochs that are respectively trained from batches $X_1,X_2$ with overlapping sample $X_c$. In this situation, dummy sample $\hat X_c$ is a common component of dummy batches $\hat X_1,\hat X_2$ and always has a single value in each dummy batch during optimization. Therefore, both updates provide information that \algo uses to optimize the sample.

If updates are observed in two or more complete epochs then all training samples can be better recovered thanks to the update matching and joint reconstruction steps. Note that an update is used $B$ times to jointly reconstruct different samples if the batch size is equal to $B$.

\begin{table}[tp]
  \centering
  \caption{PSNR results for reconstructing size-4 batches from untrained ResNet20-4, with different layer assignment functions. The decreasing functions pass through $(1, 1)$ and $(N_{conv}, 0.5)$, and the increasing functions pass through $(1, 1)$ and $(N_{conv}, 2)$, where the number of convolutional layers $N_{conv}=21$ in ResNet20-4. The two points and function expressions determine the used functions.}
    \begin{tabular}{lllcc}
    \toprule
    Monotonicity & Convexity & Function & CIFAR-10 & CIFAR-100 \\
    \midrule
    None  & None  & $y=1$ & 14.836  & 15.190  \\
    \midrule
    Decrease & Convex & $-a\log x+b$ & 14.182  & 14.332  \\
    Decrease & Concave & $-a e^x+b$ & 14.727  & 15.175  \\
    Decrease & None  & $-ax+b$ & 14.354  & 14.471  \\
    \midrule
    Increase & Convex & $a e^x+b$ & 15.255  & 16.001  \\
    Increase & Concave & $a\log x+b$ & 15.505  & 16.311  \\
    Increase & None  & $ax+b$ & \textbf{15.601}  & \textbf{16.702}  \\
    \bottomrule
    \end{tabular}%
  \label{tab:linear_shape}%
\end{table}%

\subsection{Assigning layer weights}
\label{sec:weights}

It has been observed that different layers of a neural network have variable impact on model accuracy~\cite{elkerdawy2020filter}, energy consumption~\cite{yang2017designing}, and can be selectively pruned for model compression~\cite{chen2018shallowing}. Similarly, we observe that gradients from different layers have different effect on the performance of a gradient inversion attack. 

Based on these previous works, \algo assigns different weights $\alpha_i$ to layers in the distance function used in the gradient inversion attack. For a CNN that has $N_{conv}$ convolutional layers before the fully connected layers, suppose that the input gradients can be decomposed by layers from first to last as $\nabla W = (\nabla L_1, ..., \nabla L_{N_{conv}}, \nabla L_{fc})$ and so do the reconstructed gradients $\nabla W'$. \algo's negative cosine similarity objective with layer weights $\alpha_i$ is indicated in Formula~\ref{eq:layer}.

\begin{equation} \label{eq:layer}
\min_{\hat X} 1 - \frac{\sum_i \alpha_i (\nabla L'_i \cdot\nabla L_i)}{\sqrt{\sum_i \alpha_i \parallel \nabla L'_i \parallel_2^2} \sqrt{\sum_i \alpha_i \parallel \nabla L_i \parallel_2^2}}
\end{equation}

\textbf{Linear assignment.} It is impractical to separately select a weight for each layer using tedious methods like grid search. Therefore, we use a function to assign the weights to convolutional layers, and set weights of fully connected layers to the average of convolutional layer weights. 

Both the shape and the range of the weight function affect performance. For the range, we first determine that the function passes through points $(1,1)$ and $(N_{conv}, \beta)$. Parameter $\beta$ controls the range of the function and $\beta=1$ implies homogeneous weights. We also look into various function shapes, such as linear functions $y{=}ax{+}b$, exponential-like functions $y{=}ae^x{+}b$ and logarithmic-like functions $y{=}a\log x{+}b$ to cover various function shapes (e.g., concave or convex). The two points $(1,1)$ and $(N_{conv}, \beta)$ determine values of coefficients $a,b$ in the various functions. 

Table~\ref{tab:linear_shape} presents some results obtained with different function shapes. 
We experimentally observe that an increasing linear weight function maximizes the reconstruction performance, and therefore determines a layer's weight using Formula~\ref{eq:linear}.

\begin{equation} \label{eq:linear}
l_i = 1 + \frac{\beta - 1}{N_{conv} - 1}(i - 1) 
\end{equation}

In practice, the properties of the attacked network change during the training process. For example, the magnitude of gradient values becomes smaller and the reconstruction quality deteriorates. In consequence, the best value for parameter $\beta$ changes according to the attacked training phase (as illustrated in Figure \ref{fig:exp_layer}).

{\bf ReLU modifier.} We can directly set $\alpha_i {=} l_i$ in the general case, but we additionally propose a layer weight modifier for CNNs that apply ReLU after convolution layers, as $\textnormal{ReLU}(x)=\max{(0, x)}$ is the most commonly used activation function in current CNN designs such as ResNet~\cite{he2016deep}. 
ReLU is more popular than sigmoid functions, because it simplifies gradient computation in backward propagation and avoids gradient vanishing and gradient inversion.

When the input of ReLU is negative, its output is set to zero, and as a result, after back propagation, the gradients of input are also equal to zero. Therefore, a proportion of zeros exists in each convolution layer's gradients. 
These zeros limit the performance of gradient inversion attacks, because it is then easier for the optimization process to get its dummy gradients to achieve these zeros, which appear as soon as the corresponding output of a convolutional layer is negative. Thus, layers that have more zero gradients contribute less in the reconstruction. 

To remove the effect of zeros imported by ReLU and balance the contribution of each convolutional layer, \algo relies on a layer weight modifier based on the proportion of zeros of each layer. Assuming that the proportion of zeros in a convolutional layer's gradients is $p_i$, its modifier zero-based coefficient is computed as $z_i{=}\frac{1}{1-p_i}$. 

In conclusion, the layer weight for the $i$-th convolutional layer is equal to a linearly assigned weight $l_i$, which is multiplied by a zero-based weight $z_i$ if ReLU is used, and the layer weight of a fully connected layer is the average of the linearly assigned weights of convolutional layers: 

\begin{equation} \label{eq:layer_spec}
\begin{aligned}
\alpha_{i} &= l_i \cdot \frac{1}{1-p_i} &i\textnormal{-th convolutional layer} \\
\alpha_{fc} &= \frac{\sum_{i=1}^{N_{conv}} l_i}{N_{conv}} &\textnormal{  fully connected layers}
\end{aligned}
\end{equation}

%% file: 4-Evaluation.tex
\section{Performance evaluation}
\label{sec:experiments}

\begin{table*}[tp]
	\centering
	\caption{Reconstruction quality of gradient inversion attacks on untrained ResNet20-4 on CIFAR-10 and CIFAR-100, and trained ResNet50 on ImageNet, on FedAvg model updates. Here, FedAvg runs on 4 size-1 mini-batches and 4 local steps.}
	\label{tab:exp_fedavg}
    \begin{tabular}{lcccccccrcc}
	\toprule
	Method & simulation & \multicolumn{3}{c}{CIFAR-10, untrained} & \multicolumn{3}{c}{CIFAR-100, untrained} & \multicolumn{3}{c}{ImageNet, trained} \\
	\cmidrule{3-11}          &       & PSNR\textuparrow & SSIM\textuparrow & LPIPS\textdownarrow & PSNR\textuparrow & SSIM\textuparrow & LPIPS\textdownarrow & PSNR\textuparrow & SSIM\textuparrow & LPIPS\textdownarrow \\
	\midrule
	DLG-Adam & $\checkmark$ & 13.611  & 0.391  & 0.123  & 15.114  & 0.463  & 0.096  &       8.044 & 0.028 &	1.314 \\
	InvG  & $\checkmark$ & 14.108  & 0.414  & 0.117  & 15.465  & 0.480  & 0.092  &   10.593    &     0.167  & \textbf{0.779} \\
	\midrule
	\algo (this work, one-batch) & $\times$    & 14.250  & 0.414  & 0.115  & 15.437  & 0.475  & 0.093  &  10.522     &   0.161    & 0.791 \\
	\algo (this work, one-batch + layers) & $\times$   & \textbf{16.182}      &  \textbf{0.529}     &  \textbf{0.112}     & \textbf{19.133} & \textbf{0.672} & \textbf{0.050} &   \textbf{10.699}    &  \textbf{0.177}     & 0.780 \\
	\bottomrule
\end{tabular}%
\end{table*}

\begin{table}[tp]
		\caption{Reconstruction quality of gradient inversion attacks on FedAvg, with local step 4 and batch size 2, on ResNet20-4 on CIFAR-100. The matching batches from two epochs are assumed to have the same samples. A '1' in the Epochs column means that the updates are collected after one epoch, while '1, 2' means that the matching updates are collected after one epoch and two epochs.}
	\begin{tabular}{lccccc}
		\toprule
		Method & Simulation & Epochs & PSNR\textuparrow & SSIM\textuparrow & LPIPS\textdownarrow \\
		\midrule
		DLG-Adam & $\checkmark$ & 1     & 12.196  & 0.292  & 0.087  \\
		InvG  & $\checkmark$ & 1     & 14.398  & 0.419  & 0.065  \\
		\midrule
		\algo (this work)  & $\times$    & 1     & 15.651  & 0.486  & 0.055  \\
		\algo (this work) & $\times$    & 1,2   & \textbf{16.664}  & \textbf{0.533}  & \textbf{0.040}  \\
		\bottomrule
	\end{tabular}%
	\label{tab:exp_fedavg_epoch}
\end{table}

\begin{figure}[tp]
	\centering
	\begin{subfigure}{0.45\textwidth}
		\centering
		\includegraphics[width=\linewidth, trim= 0 0 128pt 0, clip]{"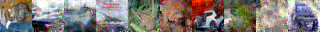"}
		\caption{DLG-Adam} 
	\end{subfigure}
	\begin{subfigure}{0.45\textwidth}
		\centering
		\includegraphics[width=\linewidth, trim= 0 0 128pt 0, clip]{"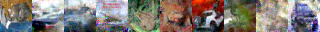"}
		\caption{InvG} 
	\end{subfigure}
	\begin{subfigure}{0.45\textwidth}
		\centering
		\includegraphics[width=\linewidth, trim= 0 0 128pt 0, clip]{"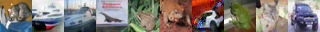"} 
		\caption{\algo (this work)} 
	\end{subfigure}
	\begin{subfigure}{0.45\textwidth}
		\centering
		\includegraphics[width=\linewidth, trim= 0 0 128pt 0, clip]{"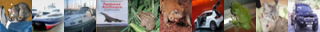"} 
		\caption{Ground truth} 
	\end{subfigure}
	\caption{Reconstructed CIFAR-10 images from untrained ResNet20-4 with batch size 1.}
	\label{fig:comp_cifar10}
\end{figure}

\begin{figure}[tp]
		\centering
	\begin{subfigure}{0.24\textwidth}
		\centering
		\includegraphics[width=0.8\linewidth]{"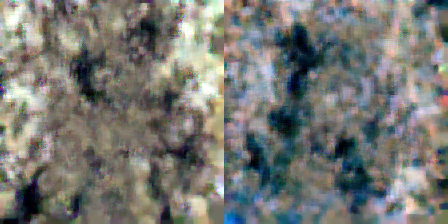"}
		\caption{DLG-Adam} 
	\end{subfigure}
	\begin{subfigure}{0.24\textwidth}
		\centering
		\includegraphics[width=0.8\linewidth]{"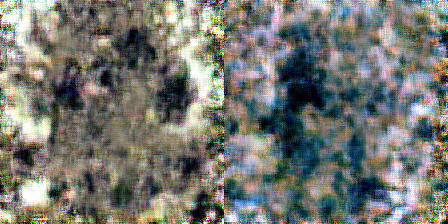"}
		\caption{InvG} 
	\end{subfigure}
	\begin{subfigure}{0.24\textwidth}
		\centering
		\includegraphics[width=0.8\linewidth]{"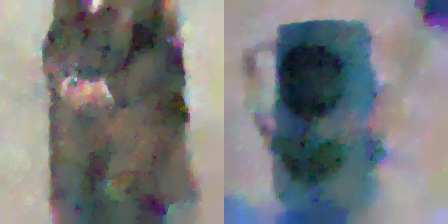"} 
		\caption{\algo (this work)} 
	\end{subfigure}
	\begin{subfigure}{0.24\textwidth}
		\centering
		\includegraphics[width=0.8\linewidth]{"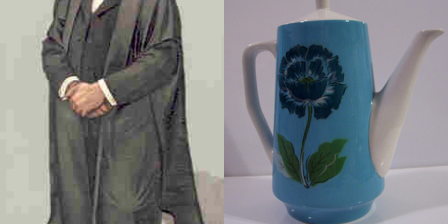"} 
		\caption{Ground truth} 
	\end{subfigure}
	\caption{Reconstructed ImageNet images from untrained ResNet50 with batch size 1.}
	\label{fig:comp_imagenet}
\end{figure}

\begin{table}[htbp]
	\centering
	\caption{Reconstruction quality of gradient inversion attacks on single gradient updates, reconstructing CIFAR-10 images from untrained ResNet20-4 and ImageNet images from untrained ResNet50.}
	\label{tab:exp_grad}
    \begin{tabular}{cp{6mm}lrrr}
	\toprule
	Dataset & Batch size & \multicolumn{1}{c}{Method} & \multicolumn{1}{l}{PSNR\textuparrow} & \multicolumn{1}{l}{SSIM\textuparrow} & \multicolumn{1}{l}{LPIPS\textdownarrow} \\
	\midrule
	\multirow{3}[2]{*}{CIFAR-10} & \multirow{3}[2]{*}{1} & DLG-Adam & 19.908  & 0.727  & 0.034  \\
	&       & InvG  & 20.671  & 0.753  & 0.031  \\
	&       & \algo (this work)  & \textbf{31.341}  & \textbf{0.963}  & \textbf{0.004}  \\
	\midrule
	\multirow{3}[2]{*}{CIFAR-10} & \multirow{3}[2]{*}{4} & DLG-Adam & 14.034  & 0.412  & 0.110  \\
	&       & InvG  & 14.421  & 0.433  & 0.103  \\
	&       & \algo (this work)  & \textbf{17.183}  &\textbf{ 0.586}  & \textbf{0.092}  \\
	\midrule
	\multirow{3}[2]{*}{ImageNet} & \multirow{3}[2]{*}{1} & DLG-Adam & 11.311  & 0.094  & 0.751  \\
	&       & InvG  & 12.832  & 0.214  & 0.691  \\
	&       & \algo (this work)  & \textbf{15.573}  & \textbf{0.366}  & \textbf{0.670}  \\
	\midrule
    \multirow{3}[2]{*}{ImageNet} & \multirow{3}[2]{*}{4} & DLG-Adam & 9.879  & 0.070  & 0.801  \\
&       & InvG  & 11.640  & 0.216  & \textbf{0.720}  \\
&       & \algo (this work) & \textbf{12.836}  & \textbf{0.302}  & 0.721  \\
\bottomrule
\end{tabular}%
\end{table}%

\subsection{Setup}\label{subsec:exp_setup}

\textbf{Hardware.} We use a machine equipped with an AMD EPYC 7542 32-core CPU, 256GB of RAM and an NVIDIA GeForce RTX 2080 Ti GPU for our experiments. 

\textbf{Datasets.} We conduct experiments on three image datasets: CIFAR-10 (size $32 \times 32$, 10 classes)~\cite{krizhevsky2009learning}, CIFAR-100 (size $32 \times 32$, 100 classes)~\cite{krizhevsky2009learning}, and ImageNet (size $224 \times 224$, 1000 classes)~\cite{russakovsky2015imagenet}. All three datasets contain 3-channel colored images. The datasets are normalized to mean $0$ and standard deviation $1$. For the experiments, we randomly select 100 mini-batches from the validation set of each dataset. We attack ResNet trained on the datasets~\cite{he2016deep}: ResNet20-4 on CIFAR-10 and CIFAR-100, and ResNet50 on ImageNet. 

\textbf{Baselines.} We compare our attack with two baselines. The first is an improved version of DLG~\cite{zhu2020deepdlg}, namely DLG-Adam, which uses the Adam optimizer instead of L-BFGS in the original version, because L-BFGS often fails to converge while attacking large networks like ResNet~\cite{geiping2020inverting}. The other baseline is Inverting Gradients~\cite{geiping2020inverting}, which we call InvG. The two baselines cover the two main distance functions applied in gradient inversion attacks.

We do not compare our attack to those based on pretrained models, e.g., an image generative model~\cite{jeon2021gradientgen}, because we do not assume the attacker to have access to additional knowledge. We also do not consider a batch normalization (BN) regularization term \cite{yin2021seegradinversion}, because it requires that each synchronous BN layer has cross-node communication during forward propagation, which is very time-consuming and impractical in FL. Our methods are however orthogonal to these methods that leverage extra information and could be combined with them. 

\textbf{Hyperparameters.} The Adam optimizer with learning rate 0.1 is used based on the gradient values on dummy samples during optimization. $\zeta_{TV}$ is equal to $10^{-4}$ for untrained ResNet20-4, $10^{-2}$ for trained ResNet20-4, $10^{-3}$ for untrained ResNet50, and $10^{-1}$ for trained ResNet50. For FedAvg, the learning rate of local SGD is $\mu=1\times 10^{-4}$. The kernel size of the averaging pooling layer used in update matching is $2\times 2$ and the stride of it is 2. Regarding epoch weights $\gamma_i$ in multi-epoch reconstruction, $\gamma=1$ for a batch from the first epoch, and $\gamma=0.1$ for the other batches. For layer weights, $\beta=50$ for untrained networks, and $\beta$=2 for trained networks. The optimization has 10,000 iterations.

Because label inference is an independent subtask that has already been well explored and for which very accurate methods already exist, we assume that labels are correctly inferred in experiments of gradient updates, if not specified in other ways. For our experiments with FedAvg, we apply the label inference method from~\cite{yin2021seegradinversion}. Labels in a mini-batch are assumed not to be duplicated, following the experiment setup of previous works~\cite{geiping2020inverting, yin2021seegradinversion}.

\textbf{Metrics.} We use three metrics to measure the similarity between the reconstructed images and the real training samples: (i) peak signal-to-noise ratio (PSNR), (ii) structural similarity index measure (SSIM)~\cite{wang2004image}, and (iii) LPIPS, a perceptual image similarity score~\cite{zhang2018unreasonable}. Higher PSNR, higher SSIM, and lower LPIPS all indicate better reconstruction quality.

\subsection{Overall performance of \algo} \label{subsec:exp_overall}

We first report the overall performance of \algo on FedAvg model updates and Gradient updates. Table~\ref{tab:exp_fedavg} shows the attack results on single FedAvg model updates. The one-batch approximation and layer weights are used with this FL setting. \algo with the one-batch approximation outperforms the other two baselines that are based on simulation of FedAvg. Table~\ref{tab:exp_fedavg_epoch} further shows the attack results on FedAvg with updates from multiple epochs. One can also observe that \algo further increases its reconstruction quality when reconstructing from matching updates from multiple epochs.

\algo can also attack gradient updates, the most common setup in gradient inversion attacks, as shown in Table~\ref{tab:exp_grad}. These results are interesting because only layer weights are used in this experiment, and \algo significantly outperforms baseline methods, with different datasets and batch sizes. \algo improves the PSNR over CIFAR-10's by up to almost  50\%. The reconstructed CIFAR-10 samples in Figure~\ref{fig:comp_cifar10} are visually almost identical to the ground truth. Figure~\ref{fig:comp_imagenet} shows two examples of reconstructed ImageNet samples, where \algo also clearly outperforms the baselines.

\begin{figure}[tp]
	\centering
\begin{subfigure}[t]{0.23\textwidth}
	\centering
	\begin{tikzpicture}[scale=0.5]
		\begin{axis}  
		[  
		ybar,
		area legend,
		enlargelimits=0.253,
		legend pos=north east, 
		x label style={font=\LARGE},
		y label style={font=\LARGE},
		ticklabel style={font=\Large},
		legend style={font=\Large},
		ylabel={PSNR}, 
		xlabel={\# local steps},
		symbolic x coords={2, 4, 8},  
		xtick=data,
		legend cell align={left},  
		]  
		\addplot coordinates {(8, 14.02) (4, 14.32) (2, 14.44)};
		\addplot coordinates {(8, 14.11) (4, 14.35) (2, 14.56)};  
		\legend{InvG, \algo (this work)}    
		\end{axis} 
	\end{tikzpicture}
	\subcaption{Reconstruction quality.}
\end{subfigure}
\begin{subfigure}[t]{0.23\textwidth}
	\centering
\begin{tikzpicture}[scale=0.5]
	\begin{axis}  
	[  
	ybar,
	area legend,
	enlargelimits=0.253,
	legend pos=north west, 
	x label style={font=\LARGE},
	y label style={font=\LARGE},
	ticklabel style={font=\Large},
	legend style={font=\Large},
	ylabel={time cost (min)}, 
	xlabel={\# local steps},
	symbolic x coords={2, 4, 8},  
	xtick=data,
	legend cell align={left},  
	]  
	\addplot coordinates {(8, 33.2) (4, 19.7) (2, 11.6)};
	\addplot coordinates {(8, 6.7) (4, 6.7) (2, 6.7)};  
	\legend{InvG, \algo (this work)}    
	\end{axis} 
\end{tikzpicture}
\subcaption{Computation time for attacks with 10k iterations.}
\end{subfigure}
\caption{Reconstruction qualities and computation times for the simulation-based attack InvG and for \algo on CIFAR-10 images from untrained ResNet20-4. Each model update is generated with 8 images, e.g., for the 8-step scenario each update is generated from 8 size-1 mini-batches.}
\label{fig:exp_bar}
\end{figure}
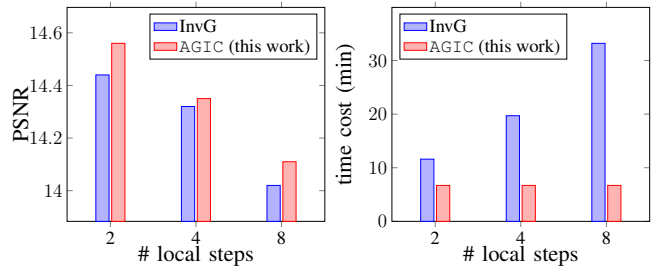

\subsection{Computation time reduction with the one-batch approximation} 
\label{subsec:exp_fedavg}

We further compare the one-batch approximation of \algo with the simulation approach of InvG, for attacking FL systems using FedAvg. We look into the case where a FedAvg update is generated from 8 images, but they are divided into mini-batches with different sizes in different experiments, which implies that the number of local steps also changes. Figure~\ref{fig:exp_bar} show that \algo outperforms InvG in both quality and speed with all settings. It can also be noticed that both methods have worse performance when the number of local steps increases, if the total number of samples are fixed. 

When the number of local steps increases, the time cost of attacks based on simulation also increases, while the one-batch approximation implies that \algo has a constant time cost. When FedAvg has 8 local steps, \algo is 5 times faster than InvG: \algo requires 6.7 minutes on an NVIDIA GeForce RTX 2080 Ti GPU instead of 33.2 minutes with InvG.

\begin{table}[tp]
	\centering
	\caption{Matching success rate of 1000 CIFAR-100 images from epochs 1 and 2, with different update matching methods and batch sizes.}
	\begin{tabular}{cccc}
		\toprule
		Labels & Pooling & Batch size 1 & Batch size 4 \\
		\midrule
		$\times$ & $\times$ & 0.819 & 0.509 \\
		$\times$ & $\checkmark$ & 0.873 & 0.547 \\
		$\checkmark$ & $\times$ & \textbf{0.988} & 0.672 \\
		$\checkmark$ & $\checkmark$ & 0.981 & \textbf{0.705} \\
		\bottomrule
	\end{tabular}%
	\label{tab:exp_gradmatch}%
\end{table}%

\begin{table}[tp]
	\centering
	\caption{Reconstruction quality with 200 CIFAR-100 images using  gradients from epochs 1 and 2, using label inference and update matching, and equal layer weights. The model is ResNet20-4 and the batch size equals 1.}
	\label{tab:exp_epoch_match1}
	\begin{tabular}{lccc}
		\toprule
		Epochs & \multicolumn{1}{c}{PSNR\textuparrow} & \multicolumn{1}{c}{SSIM\textuparrow} & \multicolumn{1}{c}{LPIPS\textdownarrow} \\
		\midrule
		1     &   20.379    &   0.665    & 0.026 \\
		2     &   19.085    &   0.626    & 0.022 \\
		1,2   &   \textbf{21.464}    &   \textbf{0.697}    & \textbf{0.016} \\
		\bottomrule
	\end{tabular}%
\end{table}%

\subsection{Reconstruction quality improvement with update matching and joint reconstruction}
\label{subsec:exp_epoch}

We now focus on reconstructing samples from multiple epochs. We first conduct experiments on our pre-reconstruction matching method with 2,000 iterations. We consider matching with and without using an average pooling layer, and with and without restriction on inferred labels. Table~\ref{tab:exp_gradmatch} presents the matching rate with pre-reconstructed CIFAR-100 images. It shows that with pre-reconstruction, the matching rate is very high (0.988) when the batch size is 1. When the batch size is 4, the matching rate decreases (0.672), because reconstruction quality decreases when the batch size increases, but it is still high. Using inferred labels provides strong prior knowledge for gradient matching, even when the batch size is larger than 1, which implies that labels may be incorrectly inferred. A pooling layer successfully mitigates noise, except in the case where the batch size equals 1 and labels are used, which already has an almost perfect accuracy.

We further evaluate on a full setup that includes label inference the matching of gradient updates and joint reconstruction. Pre-reconstruction runs 2,000 iterations. Gradient updates are matched based on pre-reconstructed single images. Table~\ref{tab:exp_epoch_match1} shows that the reconstruction results based on two updates from two epochs are better than with single updates from either epoch (e.g., PSNR of 21.464 instead of 20.379 or 19.085).

\begin{figure}[tp]
	\centering
	\begin{tikzpicture}[scale=0.6]
	\begin{semilogxaxis}
	[xmin=3e-1, xmax=1.5e2, domain=5e-1:1e2,
	ymin=10, ymax=18,
	width=0.6\textwidth,
	height=7.5cm,
	legend columns=2,
	scaled x ticks=real:1e-3,
	x label style={font=\Large},
	y label style={font=\Large},
	tick label style={font=\large},
	legend style={font=\large},
	ylabel=PSNR,
	xlabel=$\beta$: last layer weight / first layer weight,
	legend cell align={left},
	legend style={
		at={(0,1)},
		anchor=south west}],
	xtick scale label code/.code={},
	log ticks with fixed point]
	
	\addplot[smooth,mark=o,purple] plot coordinates {
		(0.5, 14.37) (1, 15.13) (2, 15.79)
		(5, 16.69) (10, 17.05) (20, 17.34)
		(50, 17.34) (100, 17.32)		
	};
	\addlegendentry{untrained, with $z_i$ from ReLU}
	
	\addplot[smooth,mark=*,blue] plot coordinates {
		(0.5, 14.35) (1, 14.83) (2, 15.60)
		(5, 16.48) (10, 16.74) (20, 17.09)
		(50, 17.22) (100, 17.16)		
	};
	\addlegendentry{untrained}
	
	\addplot[smooth,mark=*,olive] plot coordinates {
		(0.5, 12.97) (1, 13.24) (2, 13.60)
		(5, 13.59) (10, 13.74) (20, 13.80)
		(50, 13.78) (100, 13.66)		
	};
	\addlegendentry{5 epochs}
	
	\addplot[smooth,mark=*,green] plot coordinates {
		(0.5, 11.82) (1, 12.25) (2, 12.36)
		(5, 12.26) (10, 12.34) (20, 12.56)
		(50, 12.42) (100, 12.04)		
	};
	\addlegendentry{10 epochs}
	
	\addplot[smooth,mark=*,red] plot coordinates {
		(0.5, 10.86) (1, 10.99) (2, 11.31)
		(5, 10.87) (10, 10.83) (20, 11.00)
		(50, 10.90) (100, 10.64)		
	};
	\addlegendentry{trained}
	
	\addplot[thick, samples=50, dashed,domain=0:6, black] coordinates {(1,10)(1,18)};
	
	\end{semilogxaxis}
	
	\end{tikzpicture}
	\caption{Effect of different layer weights on PSNR with different experimental settings, including untrained and trained networks.}
	\label{fig:exp_layer}
\end{figure}
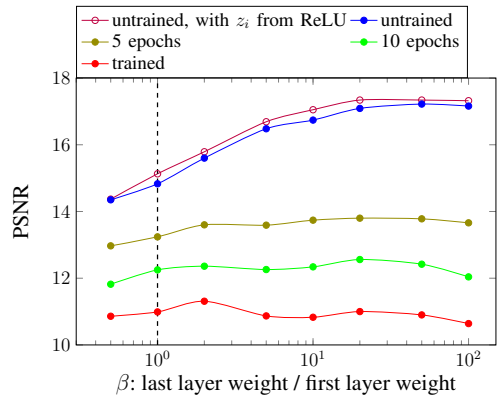

\begin{figure}[tp]
	\centering
	\includegraphics[width=0.95\linewidth]{"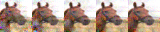"} 
	\caption{Reconstructions of a CIFAR-10 image of a horse from untrained ResNet20-4 with a weight ratio $\beta$ that ranges from 1 (left) to 20 (right).} 
	\label{fig:exp_horses}
\end{figure}

\subsection{Reconstruction quality improvement with layer weights}

We investigate the effect of layer weights on reconstruction quality by varying the ratio $\beta$ between the weights of the last convolutional layer and the first one, using ResNet20-4 on CIFAR-10. Figure~\ref{fig:exp_layer} presents our results. The effect of layer weights is very significant on the untrained network. As training continues, the effect of weights weakens, and the best $\beta$ value is also lower. However, assigning larger weights to later convolutional layers ($\beta >1$) still improves performance, compared with the equally assigned weights at $\beta=1$, even in trained networks. Figure~\ref{fig:exp_horses} shows the reconstructed images of one CIFAR-10 sample on untrained ResNet20-4, with increasing $\beta$ values. One can notice that the reconstruction quality improves when $\beta$ increases.

Figure~\ref{fig:exp_layer} also compares attacking untrained ResNet20-4 with and without ReLU weight modifier $z_i$. The zero-proportion-based weight modifier consistently improves the attack performance, bringing up to 2.0\% improvement on PSNR.

%% file: 5-related.tex
\begin{table*}[tbp]
	\centering
	\caption{Comparison of \algo and other gradient inversion attacks on (horizontal) federated learning systems. }
	\label{tab:rw_cmp}
	
    \begin{tabular}{p{11em}p{4.055em}>{\raggedright}p{8.165em}p{4.335em}p{4.055em}>{\raggedright}p{7.5em}p{3.945em}p{3.865em}p{3.70em}}
	\toprule
	Method & Main loss term & Other loss terms & Optimizer & Label inference capability & Extra knowledge & Trained network & Multiple epochs' updates & FedAvg \\
	\midrule
	DLG~\cite{zhu2020deepdlg}   & L2    & None  & L-BFGS & No use & No    & $\times$ & $\times$ & $\times$ \\
	\midrule
	iDLG~\cite{zhao2020idlg} & L2    & None  & L-BFGS & One   & No    & $\times$ & $\times$ & $\times$ \\
	\midrule
	Inverting Gradients (InvG in this paper)~\cite{geiping2020inverting} & Cosine & Total variation & Adam  & One   & No    & $\checkmark$ & $\times$ & $\checkmark$ \\
	\midrule
	CPL~\cite{wei2020framework}  & L2    & Label-based & L-BFGS & One   & No    & $\times$ & $\times$ & $\times$ \\
	\midrule
	R-GAP~\cite{zhu2020rgap} & Recursively   & N/A   & N/A   & One   & No    & $\times$ & $\times$ & $\times$ \\
	\midrule
	Grad-Inversion~\cite{yin2021seegradinversion} & L2    & Total variation, L2 of input, BN, group consistency & Adam  & Multiple & BN statistics, pretrained image alignment model & $\checkmark$ & $\times$ & $\times$ \\
	\midrule
	Geng et al.~\cite{geng2021general} & L2    & Clipping, scaling & L-BFGS & Multiple & No    & $\times$ & $\times$ & Only full batch \\
	\midrule
	GIAS \cite{jeon2021gradientgen} & Cosine & Total variation & Adam  & One   & Pretrained generative model & $\times$ & $\times$ & $\times$ \\
	\midrule
	\midrule
	\algo (this work)  & Cosine & Total variation & Adam  & Multiple & No    & $\checkmark$ & $\checkmark$ & $\checkmark$ \\
	\bottomrule
\end{tabular}%
\end{table*}%

\section{Related Work} \label{sec:relatedwork}

\subsection{Gradient inversion attack on gradient updates} \label{subsec:rw_grad}

Gradient inversion attacks on federated learning systems reconstruct training samples from gradient updates. Deep Leakage from Gradients (DLG)~\cite{zhu2020deepdlg} is the first gradient inversion attack.  
DLG jointly optimizes on the samples and on the labels, which complicates the optimization objective and impairs the reconstruction quality if the labels are inaccurately optimized. Improved DLG (iDLG)~\cite{zhao2020idlg} proposes to conduct label inference before DLG's optimization and shows that labels can be inferred with perfect accuracy using some analytical method when the batch size is equal to 1. Following iDLG, several label inference methods explained how to infer labels in larger batches with higher accuracy~\cite{yin2021seegradinversion, geng2021general, dang2021revealing}. Label inference is now an essential part of current gradient inversion attacks since it significantly improves the reconstruction quality.

Most existing gradient inversion attacks are optimization-based: they formulate an optimization problem and minimize the distance between the real observed gradient update and a dummy gradient update in order to optimize the reconstructed samples. Therefore, one way to improve the performance of a gradient inversion attack is to design a better objective function. Geiping et al. uses the cosine distance instead of L2 distance in DLG~\cite{geiping2020inverting}. CPL adds a label-based regularization term to improve the optimization stability~\cite{wei2020framework}. Another way to improve the reconstruction performance is to leverage prior knowledge in specific scenarios. GradInversion adds a batch normalization regularizer, because the attacker is assumed to know batch normalization statistics, and it designs a group consistency regularization term leveraging multiple reconstruction processes with different initialization seeds to find an enhanced reconstruction result~\cite{yin2021seegradinversion}. GIAS assumes a known data distribution and improves the reconstruction quality using a pre-trained generative model~\cite{jeon2021gradientgen}. 

R-GAP is a gradient inversion attack that is not based on optimization. Instead, R-GAP recursively reconstructs each layer's input from the last layer to first by solving linear equations, which is limited to a batch size equal to 1~\cite{zhu2020rgap}. 
The works we have discussed so far consider horizontal FL systems~\cite{kantarcioglu2004privacy, konevcny2016federated} where client datasets share a given feature space. Differently, CAFE is a gradient inversion attack~\cite{jin2021catastrophiccafe} for vertical FL systems where client datasets have different feature spaces~\cite{vaidya2002privacy, hardy2017private}.

Table~\ref{tab:rw_cmp} compares our attack, \algo, with existing gradient inversion attacks, and shows that it is applicable in more general federated learning scenarios.

\subsection{Gradient inversion attack on FedAvg model updates} \label{subsec:rw_fedavg}

Although using FedAvg model updates is more practical than using gradient updates, only a handful of gradient inversion attacks on FedAvg's model updates have been described. 

To be compatible with the multiple local steps of FedAvg, Geiping et al. simulate the local training process of multiple local steps to obtain a dummy model update, and then minimize the distance between the dummy model update and the real model update~\cite{geiping2020inverting}. However, this method is incompatible with label inference because it is unable to associate correct labels to each mini-batch, as label inference returns a set of labels for all mini-batches. In addition, the computation time of simulation-based attacks also increases with the number of local steps FedAvg uses, because the computation graph for a model update then also gets larger~\cite{geiping2020inverting}.
Geng et al. present an attack that targets FedAvg systems that use full batch gradient descent~\cite{geng2021general}. Their attack divides the received model update by the number of local steps to approximate the model update of each local step, which is then used to approximate the full batch's gradient update. However, this attack cannot cover the common case where FedAvg uses multiple mini-batches.

\algo addresses the limitations of previous attacks on FedAvg. First, its one-batch approximation is compatible with label inference. Second, \algo can be used in common FedAvg scenarios that use several mini-batches per round, and it is faster than simulation-based attacks.